\documentclass[twoside,10pt]{article}
\usepackage [top=25truemm,bottom=25truemm,left=25truemm,right=25truemm]{geometry}
\usepackage[utf8]{inputenc}
\usepackage[T1]{fontenc}
\usepackage{siunitx} 
\usepackage{graphicx} 
\usepackage{amsmath}
\usepackage{mathtools}
\usepackage[sort&compress,numbers]{natbib}
\usepackage{array}
\usepackage{diagbox}
\usepackage{multirow}  
\usepackage{tabularx}
   \newcolumntype{C}{>{\centering\arraybackslash}X}
   \newcolumntype{L}{>{\raggedright\arraybackslash}X}
   \newcolumntype{R}{>{\raggedleft\arraybackslash}X}

\title{Decentralized multi-agent reinforcement learning algorithm \\using a cluster-synchronized laser network}
\author{Shun Kotoku${}^1$, Takatomo Mihana${}^{1,*}$, Andr\'{e} R\"{o}hm${}^1$, and Ryoichi Horisaki${}^1$}
\date{}
\begin{document}
\columnseprule=0.2mm
\maketitle
\vspace{-2.1\baselineskip}
\begin{center}
{\small Department of Information Physics and Computing, Graduate School of Information Science and Technology,\\
The University of Tokyo, 7-3-1 Hongo, Bunkyo, Tokyo 113-8656, Japan.\\
$^*$Corresponding author. Email: \texttt{takatomo\_mihana@ipc.i.u-tokyo.ac.jp}
}
\end{center}

\begin{center}\textbf{Abstract}
Multi-agent reinforcement learning (MARL) studies crucial principles that are applicable to a variety of fields, including wireless networking and autonomous driving. 
We propose a photonic-based decision-making algorithm to address one of the most fundamental problems in MARL, called the competitive multi-armed bandit (CMAB) problem. 
Our numerical simulations demonstrate that chaotic oscillations and cluster synchronization of optically coupled lasers, along with our proposed decentralized coupling adjustment, efficiently balance exploration and exploitation while facilitating cooperative decision-making without explicitly sharing information among agents. 
Our study demonstrates how decentralized reinforcement learning can be achieved by exploiting complex physical processes controlled by simple algorithms. 
\end{center}\vspace{-0.5\baselineskip}

\section{\label{sec:intro} Introduction}
Reinforcement learning~\cite{Sutton1998} is a subfield of machine learning where an agent interacts with environments through trial and error, optimizing its actions based on a reward norm. 
Multi-agent reinforcement learning (MARL)~\cite{Tan1993, Stone2000, Canese2021} is an extended form of reinforcement learning where multiple agents simultaneously interact with environments as well as other agents. 
Its applications have now spread across various areas, including wireless networking~\cite{Zia2019}, autonomous driving~\cite{Shalev2016}, power distribution networks~\cite{Wang2021}, etc. 
We focus on one of the most fundamental settings of MARL: the competitive multi-armed bandit (CMAB) problem~\cite{Lai2008}. 

The CMAB problem, an extension of the multi-armed bandit (MAB) problem~\cite{Robbins1952}, features several players (agents) who repeatedly select among multiple slot machines (arms) with unknown hit probabilities (uncertain environments). 
Players need to balance two opposing strategies: exploration, which involves players attempting to identify the profitable slots by exploring various options, and exploitation, where players attempt to focus on the most promising option by exploiting the acquired information about the environments. 
The term `competitive' can be misleading in the context of MARL. 
In this setting, it does not imply a zero-sum reward distribution as it might in the literature~\cite{Canese2021}. 
Instead, the rewards are divided among players when they select the same slot, and players cooperatively aim to maximize the total rewards for the team. 
Therefore, avoiding selection collisions is important for effective decision-making. 
The CMAB problem can be considered a form of stateless MARL, provided the reward environment remains constant. 

In the past decade, several studies regarding the MAB~\cite{Naruse2014, Naruse2017, Mihana2020, Morijiri2023} or CMAB~\cite{Chauvet2019, Mihana2022, Ito2024, Kotoku2024} problem have been conducted in the context of photonic accelerators~\cite{Kitayama2019}. 
Photonic accelerators involve promising approaches to accelerate or improve performances of specific computations by utilizing photonics, such as its high propagation speed, broad bandwidth, and parallelism. 
Our study features a cooperative decision-making system using a laser network~\cite{Ito2024, Kotoku2024}, which leverages the rapid and chaotic oscillations and synchronization observed in optically interconnected semiconductor lasers for avoiding slot selection collisions in the CMAB problem. 

Specifically, the laser-based decision-making system employs two kinds of coexisting laser interactions: a leader-laggard relationship and cluster synchronization. 
The leader-laggard relationship~\cite{Heil2001, Kanno2017} observed in coupled lasers with chaotic dynamics is similar to delay-synchronization, where the temporal waveform of the `leader' laser's intensity precedes that of the `laggard' laser by the coupling delay time. 
Unlike full delay-synchronization commonly observed in delay-coupled networks, the roles of the leader and laggard spontaneously exchange. 
Cluster synchronization~\cite{Ingo2006, Nixon2011, Dahms2012, Pecora2014} refers to a state where nodes in a delay-coupled network separate into several clusters, and the oscillations of nodes within the same cluster synchronize without delay. 

The decision-making system assigns a subset of lasers in a network to players. 
Each laser corresponds to each player's selection of slot machines, and players select slots based on the leader laser at each decision point. 
Cluster synchronization in a laser network enables players to avoid selection collisions. 
The previous study~\cite{Kotoku2024} has revealed the balance of the leader-laggard relationship can be controlled while maintaining two-cluster synchronization by adjusting coupling strengths between lasers in four laser networks, corresponding to the CMAB problem with two players and two slots. 

However, the previous work~\cite{Kotoku2024} has been limited to investigating laser dynamics and interactions with parameters fixed for each setting, and it has not addressed the CMAB problem itself. 
Specifically, it has not yet been discussed how coupling strengths should be dynamically adjusted. 
An earlier approach~\cite{Mihana2022} has also addressed photonic-based collective decision-making, but the method presumed the existence of a centralized controller besides the players. 
Physics-based methods, and especially photonic approaches, for decision-making are likely to find applications in edge-devices, which lack the local computing or networking resources to implement traditional digital learning algorithms.
In the case of such edge-devices, it is more realistic and practical to assume that each player can only observe the result of its own slot selection. 
In addition, the previous work~\cite{Kotoku2024} has dealt only with four-laser networks for the two-player and two-slot CMAB problem. 
Selection only truly matters, when the number of arms exceeds the number of players, as otherwise avoiding selection collisions is the only important aspect of maximizing rewards for the team. 

In this study, we propose an enlarged laser network that exhibits stable cluster synchronization, as well as an algorithm to adjust the coupling strengths of the laser network based on the player's individually received rewards to efficiently address the competitive multi-armed bandit (CMAB) problem with more slot machines than players. 
Our numerical simulations validate the viability of our proposed laser network and decentralized coupling adjustment algorithm for tackling the CMAB problem. 

Note that we do not assert that the proposed photonic-based decision-making algorithm is superior to existing methods, such as the Upper Confidence Bound 1 (UCB1)~\cite{Auer2002} for the single-agent configuration and its variants for the multi-agent settings, in all aspects and can entirely replace them. 
Rather, we focus on exploring the application of nontrivial photonics phenomena accompanied by straightforward and intuitive learning algorithms.
Our results show that the proposed system is indeed reaching good performance in the explored circumstances. 

\section{\label{sec:method} Method and dynamics investigation}
\subsection{\label{subsec:system} System configuration}
We study the case of the competitive multi-armed bandit (CMAB) problem with two players and three slot machines. 
\renewcommand{\arraystretch}{1.5}
\begin{table}[b]
\caption{An example of a payoff matrix for the rewards in a competitive multi-armed bandit problem with two players and three slot machines ($P_{\mathrm{A}} = 0.4, P_{\mathrm{B}} = 0.6, P_{\mathrm{C}} = 0.6$).
The first and second elements represent the expected rewards for Players 1 and 2.
The underlined portions along the diagonal indicate where selection collisions happen. }
\label{tab:problem}
\begin{center}
{
 \begin{tabularx}{0.5\linewidth}{|>{\centering}p{2.1em}|C|C|C|} 
 \hline 
  \diagbox[height=1.8\line]{2}{1} & Slot A & Slot B & Slot C \\
 \hline
  A & (\underline{0.2, 0.2}) & (0.6, 0.4) & (0.6, 0.4)\\
  \hline
  B & (0.4, 0.6) & (\underline{0.3, 0.3}) & (0.6, 0.6) \\
  \hline
  C & (0.4, 0.6) & (0.6, 0.6) & (\underline{0.3, 0.3})\\
 \hline 
 \end{tabularx}
 }
\end{center}
\end{table}
Table~\ref{tab:problem} shows an example of a payoff matrix for the rewards in the CMAB problem with a typical problem setting of the hit probabilities of Slots~A, B, and C: $P_{\mathrm{A}} = 0.4, P_{\mathrm{B}} = 0.6, P_{\mathrm{C}} = 0.6$. 
We assume a temporally static reward environment for simplicity. 
Each column in Table~\ref{tab:problem} represents the slot selected by Player 1, and each row represents the slot by Player 2. 
The underlined portions along the diagonal indicate cases where the players select the same slot, and the rewards are thus divided among them. 
Due to this division of spoils, the team's sum rewards are lower in such cases compared to other selection patterns. 
Naturally, players are encouraged to focus on selecting the most high-paying two out of three slots in the exploitation phase after sufficient exploration to estimate the hit probabilities of the slots. 
Thus, it is optimal to narrow down their selections to Slots B and C, which yield the maximum potential rewards in the problem setting. 
But at the same time, they need to avoid both selecting the same slot.

\begin{figure}[t]
\centering\includegraphics[width = 0.5\linewidth]{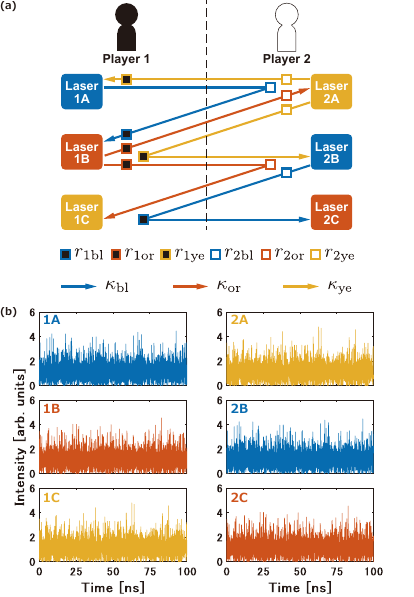}
\caption{Schematic illustration of our proposed system. (a) A six-laser network to address the competitive multi-armed bandit problem with two players and three slot machines.  
$r_{1\mathrm{\sharp}}$ and $r_{2\mathrm{\sharp}}$ $(\mathrm{\sharp} = \mathrm{bl}, \mathrm{or}, \mathrm{ye})$ represent the attenuation rates adjusted by Players 1 and 2. 
$\kappa_{\mathrm{\sharp}}$ represents the total multiplicative coupling strength. (b) Typical laser intensity waveforms of the six-laser network obtained through numerical simulations. 
Lasers drawn with identical colors are synchronized.}
\label{fig:schematic}
\end{figure}
Figure~\ref{fig:schematic}~(a) illustrates the proposed six-laser network to address the CMAB problem with a two-player and three-slot-machine configuration. 
Note that the coupling delay time between lasers, $\tau$, is assumed to be uniform for all connections.  
Lasers 1A, 1B, and 1C are allocated to Player 1, and Lasers 2A, 2B, and 2C to Player 2. 
Each laser corresponds to a slot machine selected by a player. 

In the collective decision-making system, players select a slot corresponding to the leader among the three lasers assigned to them. 
The leader is defined by short-term cross-correlation (STCC) values in the following formulas~\cite{Mihana2020}: 
\begin{align}
\label{eq:stcc1}
C_{\rm{1A}}(t) &= \int_{t-\tau}^{t} \frac{I_{\rm{1A}}(u) - \bar{I}_{\rm{1A}}}{\sigma_{\rm{1A}}} \frac{I_{\rm{1B}}(u - \tau) - \bar{I}_{\rm{1B},\tau}}{\sigma_{\rm{1B},\tau}} du,\\
\label{eq:stcc2}
C_{\rm{1B}}(t) &= \int_{t-\tau}^{t} \frac{I_{\rm{1B}}(u) - \bar{I}_{\rm{1B}}}{\sigma_{\rm{1B}}} \frac{I_{\rm{1C}}(u - \tau) - \bar{I}_{\rm{1C},\tau}}{\sigma_{\rm{1C},\tau}} du,\\
\label{eq:stcc3}
C_{\rm{1C}}(t) &= \int_{t-\tau}^{t} \frac{I_{\rm{1C}}(u) - \bar{I}_{\rm{1C}}}{\sigma_{\rm{1C}}} \frac{I_{\rm{1A}}(u - \tau) - \bar{I}_{\rm{1A},\tau}}{\sigma_{\rm{1A},\tau}} du,
\end{align}
where $I_k(t)$ denotes the laser intensity of Laser $k$, $\bar{I}_k$ and $\sigma_k$ represent the average and the standard deviation of $I_k(t)$, over the period of $\tau$. 
$\bar{I}_{k,\tau}$ and $\sigma_{k,\tau}$ have comparable meanings, but the time window for the calculation is shifted back by $\tau$. 
Player 1 considers the laser with the smallest STCC value of the three as the leader. 
Therefore, for example, Player 1 selects Slot B when $C_{\mathrm{1B}} < C_{\mathrm{1A}}$ and $C_{\mathrm{1B}} < C_{\mathrm{1C}}$ hold. 
STCC values for Player 2, $C_{\mathrm{2A}}, C_{\mathrm{2B}}$, and $C_{\mathrm{2C}}$, are defined similarly to Eqs.~\eqref{eq:stcc1}--\eqref{eq:stcc3}. 

In the six-laser network depicted in Fig.~\ref{fig:schematic}~(a), lasers with the same color, Lasers 1A and 2B, Lasers 1B and 2C, and Lasers 1C and 2A, are in perfect zero-lag synchronization when the coupling strengths with the same color are uniform, as shown in Fig.~\ref{fig:schematic}~(b). 
Here, we define cluster bl (blue) to consist of Lasers 1A and 2B, cluster or (orange) to consist of Lasers 1B and 2C, and cluster ye (yellow) to consist of Lasers 1C and 2A. 
Note that this color differentiation is just for convenience and is not related to the wavelength of the lasers. 
Players avoid selection collisions without sharing information about their slot selections and resultant rewards thanks to this three-cluster synchronization. 
For instance, when Laser 1B is regarded as the leader and Player 1 selects Slot B, due to the cluster synchronization with Laser 1B, Laser 2C is considered the leader by Player 2, who will then select Slot C. 
Thus, Players 1 and 2 always select different slot machines and achieve conflict avoidance in the CMAB problem. 

\subsection{\label{subsec:Leader}Leader probability and cluster synchronization}
Our previous work~\cite{Kotoku2024} has revealed that the leader-laggard relationship, which corresponds to the selection of slot machines, can be controlled by adjusting the coupling strengths while maintaining two-cluster synchronization in a four-laser network. 
Here, we numerically examine how the leader-laggard relationship in a six-laser network, as illustrated in Fig.~\ref{fig:schematic}~(a), can be controlled with coupling strengths. 

The model of chaotic semiconductor lasers is described by Lang-Kobayashi equations~\cite{Lang1980} as follows: 
\begin{align}
    \label{eq:langkobaE}
    \frac{dE_k(t)}{dt} &= \frac{1 + i\alpha}{2}\left[\frac{G_N[N_k(t) - N_0]}{1 + \varepsilon|E_k(t)|^2}-\frac{1}{\tau_p}\right]E_k(t) + \displaystyle\sum\kappa_{l\rightarrow k}E_l(t-\tau)\exp{(-i\omega\tau)}, \\
    \label{eq:langkobaN}
    \frac{dN_k(t)}{dt} &= J - \frac{N_k(t)}{\tau_s} - \frac{G_N[N_k(t) - N_0]}{1 + \varepsilon|E_k(t)|^2}|E_k(t)|^2,
\end{align}
where $E_k(t)$ and $N_k(t)$ represent the complex electric field and the carrier density of Laser $k$ at time $t$. 
The laser intensity is the square of the absolute value of the complex electric field: $I_k(t) = |E_k(t)|^2$. 
$\omega$ represents the angular frequency of the lasers. 
As a prerequisite for cluster synchronization, the laser frequencies within the same cluster should be equal. 
For simplicity, we assume identical frequencies for all lasers. 
$\kappa_{l\rightarrow k}$ is the coupling strength from Laser $l$ to Laser $k$, where the value is zero between unconnected lasers. 

\begin{table}[t]
\caption{Parameters of the Lang-Kobayashi equations. }
\label{tab:param}
\begin{center}
{
\footnotesize
     \begin{tabular}{ccc}
         \hline
         Symbol & Parameter & Value  \\
         \hline
         $G_N$ & Gain coefficient & \SI{8.40E-13}{\metre\cubed\per\second}\\
         $N_0$ & Carrier density at transparency & \SI{1.40E24}{\per\metre\cubed}\\
         $\varepsilon$ & Gain saturation coefficient & \num{2.0E-23}\\
         $\tau_p$ & Photon lifetime & \SI{1.927E-12}{\second}\\
         $\tau_s$ & Carrier lifetime & \SI{2.04E-9}{\second}\\
         $\alpha$ & Linewidth enhancement factor & \num{3.0}\\
         $\tau$ & Coupling delay time of light & \SI{5.0E-9}{\second}\\
         $J/J_{th}$ & Normalized injection current & \num{2.0}\\
         $\lambda$ & Optical wavelength of lasers & \SI{1.537E-6}{\metre}\\
         \hline
 \end{tabular}
 }
\end{center}
\end{table}
The parameters are shown in Table~\ref{tab:param} and are taken from references~\cite{Mihana2020, Ito2024, Kotoku2024} except for the injection current $J$. 
Previous studies set the value of $J/J_{th}$ to 1.1 ($J_{th}$ is injection current threshold) so that the lasers exhibit low-frequency fluctuation dynamics. 
However, according to our current examination, the setting of $J/J_{th} = 2.0$ seems more promising for stable cluster synchronization and rapid decision-making. 

The necessary conditions for cluster synchronization in the six-laser network are $\kappa_{\mathrm{1A}\rightarrow\mathrm{1B}} = \kappa_{\mathrm{2B}\rightarrow\mathrm{2C}}$ $(\equiv \kappa_{\mathrm{bl}})$, $\kappa_{\mathrm{1B}\rightarrow\mathrm{1C}} = \kappa_{\mathrm{1B}\rightarrow\mathrm{2A}}$ $(\equiv \kappa_{\mathrm{or}})$, and $\kappa_{\mathrm{2A}\rightarrow\mathrm{1A}} = \kappa_{\mathrm{2A}\rightarrow\mathrm{2B}}$ $(\equiv \kappa_{\mathrm{ye}})$. 
We conduct the numerical simulations to obtain temporal waveforms of laser intensity $I_k(t)$, with the balance of coupling strengths varied while maintaining the requirements. 
Afterward, STCC values are calculated to determine the leader-laggard relationship for each setting of coupling strengths. 

\begin{figure}[t]
\centering\includegraphics[width = 0.5\linewidth]{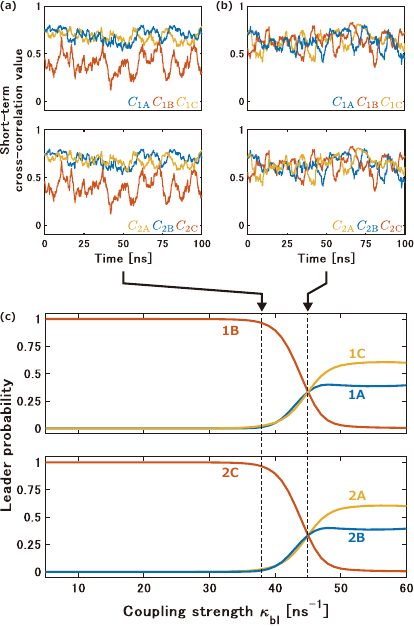}
\caption{Numerical simulation results to investigate the leader probabilities of the six-laser network shown in Fig.~\ref{fig:schematic}. (a) Short-term cross-correlation (STCC) waveforms calculated for coupling strength $\kappa_{\mathrm{or}} = \kappa_{\mathrm{ye}} = \SI{45}{\per\nano\second}$ and $\kappa_{\mathrm{bl}} = \SI{38}{\per\nano\second}$. (b) $\kappa_{\mathrm{bl}} = \SI{45}{\per\nano\second}$. (c) The relationship between the coupling strengths and leader probabilities. $\kappa_{\mathrm{or}}$ and $\kappa_{\mathrm{ye}}$ are fixed at \SI{45}{\per\nano\second}, and $\kappa_{\mathrm{bl}}$ is set from \SIrange{5}{60}{\per\nano\second}. }
\label{fig:wfstcc}
\end{figure}

Figure~\ref{fig:wfstcc} (a) and (b) show STCC time series acquired from numerical simulations. 
Coupling strengths are configured as $\kappa_{\mathrm{or}} = \kappa_{\mathrm{ye}} = \SI{45}{\per\nano\second}$ and (a) $\kappa_{\mathrm{bl}} = \SI{38}{\per\nano\second}$, and (b) $\kappa_{\mathrm{bl}} = \SI{45}{\per\nano\second}$. 
Perfect three-cluster synchronization is observed for both settings. 
$C_{\mathrm{1B}}$ and $C_{\mathrm{2C}}$ are the smallest in most cases for (a), while the frequent switching among $C_{\mathrm{1A}}$, $C_{\mathrm{1B}}$, and $C_{\mathrm{1C}}$ and that among $C_{\mathrm{2A}}$, $C_{\mathrm{2B}}$, and $C_{\mathrm{2C}}$ are observed for the symmetric setting of coupling strengths, (b). 

We define the leader probability as the ratio of times each laser is identified as a leader. 
The leader probabilities are calculated from STCC waveforms spanning \SI{10000}{\nano\second} after waiting for a sufficiently long transient of \SI{3000}{\nano\second}. 
We repeat the computation 50 times with the initial states of the lasers randomized each time and calculate the average leader probabilities of lasers for each $\kappa_{\mathrm{bl}}$. 
$\kappa_{\mathrm{or}}$ and $\kappa_{\mathrm{ye}}$ are fixed at \SI{45}{\per\nano\second}, and $\kappa_{\mathrm{bl}}$ is varied from \SIrange{5}{60}{\per\nano\second}, in increments of \SI{1}{\per\nano\second}. 
The average leader probabilities of the six lasers in the network are shown in Fig.~\ref{fig:wfstcc}~(c). 
The leader probabilities of 1A, 1B, and 1C (2A, 2B, and 2C) are approximately one-third for the symmetric coupling strengths, $\kappa_{\mathrm{bl}} = \SI{45}{\per\nano\second}$. 
The leader probability of 1B (2C) is greater than one-third with $\kappa_{\mathrm{bl}}$ smaller than \SI{45}{\per\nano\second}, and converges toward \num{1.0} around $\kappa_{\mathrm{bl}} = \SI{30}{\per\nano\second}$. 
An attempt to reduce the value of $\kappa_{\mathrm{bl}}$ smaller than \SI{5}{\per\nano\second} brings about the cessation of the chaotic oscillations in the laser network, and we want to focus on the region in which the chaotic leader-laggard relationship and cluster synchronization coexist. 
Conversely, the leader probability of 1B (2C) is smaller than one-third for $\kappa_{\mathrm{bl}} > \SI{45}{\per\nano\second}$, and converges toward 0 around $\kappa_{\mathrm{bl}} = \SI{60}{\per\nano\second}$. 

Note that similar results are obtained when $\kappa_{\mathrm{ye}}$ and $\kappa_{\mathrm{bl}}$ ($\kappa_{\mathrm{bl}}$ and $\kappa_{\mathrm{or}}$) are fixed and $\kappa_{\mathrm{or}}$ ($\kappa_{\mathrm{ye}}$) is varied because of the symmetry of the network. 

Therefore, when players estimate the combination of Slots B and C is optimal after the exploration phase, they are encouraged to decrease $\kappa_{\mathrm{bl}}$ in the exploitation phase so that Lasers 1B and 2C are more likely to be leaders. 
Similarly, if the option of Slots C and A is good, players lower $\kappa_{\mathrm{or}}$, and if that of Slots A and B is good, they lower $\kappa_{\mathrm{ye}}$. 

\subsection{\label{subsec:tow} Decentralized coupling adjustment}

The decision-making process in our proposed system is as follows. 
First, two players observe the laser intensities and calculate short-term cross-correlation (STCC) values to determine the leader of the three lasers allocated to them. 
Players select the slot corresponding to the leader, i.e., the one with the lowest STCC value. 
In perfect cluster synchronization, players naturally and independently select different slots. 
Based on the resulting slot rewards, players adjust the optical attenuation rates, represented by $r_{\mathrm{1\sharp}}$ and $r_{\mathrm{2\sharp}}$ $(\mathrm{\sharp} = \mathrm{bl}, \mathrm{or}, \mathrm{ye})$ in Fig.~\ref{fig:schematic}~(a), to improve their subsequent slot selections, and to narrow down to the most high-paying two slots. 
Hereinafter, the whole process is referred to as one `Play,' and players attempt to maximize the team's accumulated rewards over a significant number of Plays. 

The total multiplicative coupling strength $\kappa_{\mathrm{\sharp}}$ is denoted as $\kappa_{\mathrm{\sharp}} = r_{\mathrm{1\sharp}}r_{\mathrm{2\sharp}}\kappa$, where $\kappa$ represents the coupling strength without amplification or attenuation. 
The value of $\kappa$ is determined by parameters of Lang-Kobayashi equations, and $\kappa = \SI{155.3}{\per\nano\second}$ in this paper. 

Therefore, the remaining challenge is to design an algorithm, for how players should adjust their optical attenuation rates.
With the goal of being usable in edge-computing scenarios, we designed the algorithm to be simple and robust.
The underlying idea is inspired by tug-of-war (TOW) dynamics~\cite{Kim2015}, originally designed for the multi-armed bandit (MAB) problem with a single player. 
Although several algorithms have been proposed to solve the multi-agent multi-armed bandit problem, including the Upper Confidence Bound 1 (UCB1)-based methods~\cite{Liu2010, Kalathil2014, Landgren2021}, the TOW-based algorithm effectively leverages the probabilistic property observed in photonics, and it has been applied for addressing the MAB problem with a single player using chaotic lasers~\cite{Naruse2017}. 

A modified version of the TOW-based algorithm for the CMAB problem with multiple players has been proposed later~\cite{Kim2016}, and it has been applied for the software-based conflict avoidance principle in decision-making using laser networks~\cite{Mihana2022}. 
However, the method requires players to share information about the slot rewards, which we consider inappropriate given the partial observability assumptions of the CMAB problem. 

Our proposed method, decentralized coupling adjustment (DCA), is described by the following formulas: 
\begin{align}
\label{eq:reltow1}
Q_{1,\mathrm{X}} &= 2 \bar{P}_{1, \mathrm{X}} - (\bar{P}_{\mathrm{1,2nd}} + \bar{P}_{\mathrm{1,3rd}}) , 
\end{align}
\vspace{-20pt}
\begin{align}
\label{eq:reltow3}
r_{1\sharp} = 
\begin{cases}
         r_{\mathrm{low}} &(r_{\mathrm{ini}} + r_{\mathrm{step}} Q_{1,\mathrm{S}(\mathrm{\sharp})} < r_{\mathrm{low}}),\\
         r_{\mathrm{upp}} &(r_{\mathrm{upp}} < r_{\mathrm{ini}} + r_{\mathrm{step}} Q_{1,\mathrm{S}(\mathrm{\sharp})}),\\
         r_{\mathrm{ini}} + r_{\mathrm{step}} Q_{1,\mathrm{S}(\mathrm{\sharp})}&(\text{otherwise}).
    \end{cases} 
\end{align}
$Q_{1,\mathrm{X}}$ represents the excess hit probability of Slot X $(\mathrm{X} = \mathrm{A}, \mathrm{B}, \mathrm{C})$ over the baseline measured by Player 1, where $\bar{P}_{1, \mathrm{X}}$ is the observed hit probability of Slot X by Player 1, and $\bar{P}_{1, \mathrm{2nd}}$ and $\bar{P}_{1, \mathrm{3rd}}$ denote the observed hit probability of the second and third-best slot machine for Player 1. 
The excess hit probability $Q_{1,\mathrm{X}}$ of the best and second slot becomes positive, whereas that of the third (worst) slot becomes negative after a sufficient number of Plays. 

To keep the coupling strengths $\kappa_{\sharp} = r_{1\sharp}r_{2\sharp}\kappa$ within a certain range, $[\kappa_{\mathrm{low}}, \kappa_{\mathrm{upp}}]$, lower and upper limits of the attenuation rates $r_{\mathrm{low}} \equiv \sqrt{\kappa_{\mathrm{low}}/\kappa}$ and $r_{\mathrm{upp}} \equiv \sqrt{\kappa_{\mathrm{upp}}/\kappa}$ are established. 
$\kappa_{\mathrm{low}}$ and $\kappa_{\mathrm{upp}}$ are hyperparameters that determine the balance between exploration and exploitation. 
$r_{\mathrm{ini}} = (r_{\mathrm{low}} + r_{\mathrm{upp}})/2$ is the baseline attenuation rate. 
$r_{\mathrm{step}}$ is a scaling factor that affects the strengths of exploitation. 
$\mathrm{S}(\sharp)$ is a mapping from coloring to slot machines: $\mathrm{S}(\mathrm{bl}) = \mathrm{A}, \mathrm{S}(\mathrm{or}) = \mathrm{B}, \mathrm{S}(\mathrm{ye}) = \mathrm{C}$. 

The excess hit probability of Slot X observed by Player 2, $Q_{2, \mathrm{X}}$, is defined similarly to Eq.~\eqref{eq:reltow1}, and Player 2 adjusts the attenuation rate $r_{2\sharp}$ in the same manner as Eq.~\eqref{eq:reltow3}. 

With the DCA method, players control the leader probabilities of lasers and narrow down their slot selections. 
For example, under the problem setting shown in Table~\ref{tab:problem}, when Slot A appears low-rewarding to Player 1, $Q_{1,\mathrm{A}}$ decreases, and $r_{1\mathrm{bl}} = r_{\mathrm{ini}} + r_{\mathrm{step}}Q_{1,\mathrm{S}(\mathrm{bl})} = r_{\mathrm{ini}} + r_{\mathrm{step}}Q_{1,\mathrm{A}}$ is reduced. 
Player 2 also lowers $r_{2\mathrm{bl}}$, and thus, $\kappa_{\mathrm{bl}} = r_{1\mathrm{bl}}r_{2\mathrm{bl}}\kappa$ is reduced. 
As shown in Fig.~\ref{fig:wfstcc}~(c), when $\kappa_{\mathrm{bl}}$ is smaller than $\kappa_{\mathrm{or}}$ and $\kappa_{\mathrm{ye}}$, the leader probabilities of Lasers 1B and 2C becomes higher. 
Therefore, players can correctly select Slots B and C at a high rate. 
Notably, players share no information about observed hit probabilities of slots during decision-making in our proposed algorithm, which sets it apart from previous studies~\cite{Kim2016, Mihana2022}. 

\section{\label{sec:result} Decision-making simulations}
\subsection{\label{subsec:onetry} Results of a single trial}
We numerically solve the competitive multi-armed bandit (CMAB) problem with our decision-making algorithm to prove its effectiveness in balancing exploration and exploitation while avoiding selection conflicts. 
First, we present a sample result of a single decision-making trial in this section. 

\begin{table}[b]
\caption{Parameters of the competitive multi-armed bandit (CMAB) problem and decentralized coupling adjustment (DCA) used in Sec.~\ref{subsec:onetry}. }
\label{tab:param2}
\begin{center}
{
\footnotesize
     \begin{tabular}{ccc}
         \hline
         Symbol & Parameter & Value  \\
         \hline
         $(P_{\mathrm{A}}, P_{\mathrm{B}}, P_{\mathrm{C}})$ & Hit probabilities of slots & (0.4, 0.6, 0.6)\\
         $r_{\mathrm{step}}$ & Scaling factor & \num{1.0}\\
         $\kappa_{\mathrm{low}}$ & Lower bound of coupling strength & \SI{38}{\per\nano\second}\\
         $\kappa_{\mathrm{upp}}$ & Upper bound of coupling strength & \SI{45}{\per\nano\second}\\
         \hline
 \end{tabular}
 }
\end{center}
\end{table}
To clearly illustrate the transition from exploration to exploitation, which is the essential aspect of the MAB problem, we employ the configurations of parameters of the CMAB problem and decentralized coupling adjustment (DCA) shown in Table~\ref{tab:param2} as control conditions. 
Hit probabilities of slot machines are the same as the setting shown in Table~\ref{tab:problem}. 
Therefore, selecting Slots B and C is an optimal and correct option. 
Scaling factor $r_{\mathrm{step}}$ and lower and upper bound of coupling strengths $\kappa_{\mathrm{low}}$ and $\kappa_{\mathrm{upp}}$ are the hyperparameters of the DCA method, whose influence will be discussed in detail in Sec.~\ref{subsec:kappa}.

The decision-making interval is set to \SI{1.0}{\nano\second}. 
Our preliminary analysis indicates that the decision-making frequency should be comparable to or slower than the coupling delay time $\tau$. 
Otherwise, consecutive slot selections are positively correlated, leading to ineffective exploration and more Plays required for appropriate exploitation. 

\begin{figure}[tb]
\centering\includegraphics[width = 0.4\linewidth]{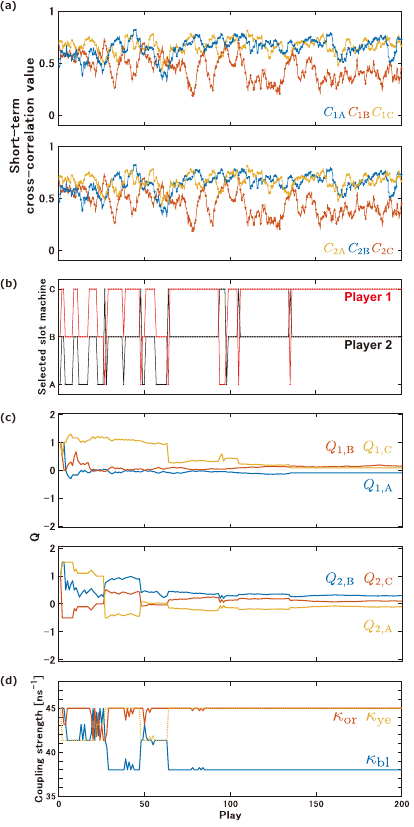}
\caption{Numerical simulation results of a single trial of decision-making. (a) Short-term cross-correlation (STCC). (b) Slot machines selected by Player 1 (red) and Player 2 (black). (c) The excess hit probabilities of slots $Q_{1, \mathrm{X}}$ and $Q_{2, \mathrm{X}}$ $(\mathrm{X} = \mathrm{A}, \mathrm{B}, \mathrm{C})$. (d) Total coupling strengths $\kappa_{\sharp} = r_{1\sharp}r_{2\sharp}\kappa$ $(\mathrm{\sharp} = \mathrm{bl}, \mathrm{or}, \mathrm{ye})$. }
\label{fig:dmstcc}
\end{figure}
Figure~\ref{fig:dmstcc} presents the numerical simulation results of a single decision-making trial. 
Figure~\ref{fig:dmstcc}~(a) shows the short-term cross-correlation (STCC) values. 
Overall, the waveforms of $C_{\mathrm{1A}}$ and $C_{\mathrm{2B}}$, $C_{\mathrm{1B}}$ and $C_{\mathrm{2C}}$, and $C_{\mathrm{1C}}$ and $C_{\mathrm{2A}}$ remain synchronized due to the consistent three-cluster synchronization. 
Therefore, as shown in Fig.~\ref{fig:dmstcc}~(b), Players 1 and 2 always select different slots, completely avoiding conflicts. 

In the first approximately 60~Plays, the relative order of $C_{\mathrm{1A}}$, $C_{\mathrm{1B}}$, and $C_{\mathrm{1C}}$, and that of $C_{\mathrm{2A}}$, $C_{\mathrm{2B}}$, and $C_{\mathrm{2C}}$, frequently change, as shown in Fig.~\ref{fig:dmstcc}~(a). 
Hence, players explore three options in turn, as shown in Fig.~\ref{fig:dmstcc}~(b).  
The fluctuations in the excess hit probabilities of slots $Q_{1, \mathrm{X}}$ and $Q_{2, \mathrm{X}}$ $(\mathrm{X} = \mathrm{A}, \mathrm{B}, \mathrm{C})$ in response to the actual received rewards are shown in Fig.~\ref{fig:dmstcc}~(c). 
Based on these $Q_{1, \mathrm{X}}$ and $Q_{2, \mathrm{X}}$, Players 1 and 2 adjust the attenuation rates $r_{1\sharp}$ and $r_{2\sharp}$ $(\mathrm{\sharp} = \mathrm{bl}, \mathrm{or}, \mathrm{ye})$ according to DCA, and the resulting coupling strengths $\kappa_{\sharp} = r_{1\sharp}r_{2\sharp}\kappa$ fluctuate in the exploration phase, as shown in Fig.~\ref{fig:dmstcc}~(d). 

Subsequently, after the exploration of roughly 60~Plays, $C_{\mathrm{1B}}$ and $C_{\mathrm{2C}}$ are the smallest in most cases, indicating that Lasers 1B and 2C are leading.
Thus, players primarily select the optimal option, Slots B and C, as shown in Fig.~\ref{fig:dmstcc}~(a) and (b). 
As expected, the excess hit probabilities for the best and second slot, $Q_{1, \mathrm{B}}$, $Q_{1, \mathrm{C}}$, $Q_{2, \mathrm{B}}$, and $Q_{2, \mathrm{C}}$, become positive, while those for the worst slot, $Q_{1, \mathrm{A}}$ and $Q_{2, \mathrm{A}}$, become negative (see Fig.~\ref{fig:dmstcc}~(c) after around 60~Plays). 
Therefore, as shown in Fig.~\ref{fig:dmstcc}~(d), $\kappa_{\mathrm{bl}}$ converges to $\kappa_{\mathrm{low}} = \SI{38}{\per\nano\second}$ whereas $\kappa_{\mathrm{or}}$ and $\kappa_{\mathrm{ye}}$ converge to $\kappa_{\mathrm{upp}} = \SI{45}{\per\nano\second}$, resulting in the appropriate focused selections of slots. 

Thus, in the case of $(P_{\mathrm{A}}, P_{\mathrm{B}}, P_{\mathrm{C}}) = (0.4, 0.6, 0.6)$, our proposed algorithm operates as intended and can successfully solve the CMAB problem. 
However, it is necessary to evaluate the decision-making performance under various problem scenarios to demonstrate the robustness of the system. 

\subsection{\label{subsec:reward} Impact of reward distributions}
In this section, we investigate the behavior of our proposed decentralized coupling adjustment (DCA) under various reward distributions to validate its robustness. 
For simplicity, we assume henceforth that the optimal combination remains Slots B and C. 
Considering the probabilistic nature of both the chaotic-laser-based decision-making system and the rewards from slot machines in the competitive multi-armed bandit (CMAB) problem, performance should be assessed as an average over multiple trials. 

We introduce the correct decision ratio (CDR)~\cite{Naruse2014} and regret as performance metrics. 
$\mathrm{CDR}(m)$ is defined by the ratio at which players correctly select the most profitable set of slot machines at $m$-th Play over a predefined number of trials. 
Each trial consists of 1000~Plays of decision-making, and this process is repeated for 2000 trials, with the six lasers randomly initialized each time. 
A faster convergence of CDR to a value closer to 1.0 indicates a better performance. 
Regret is the difference between the expected cumulative reward for the perfect selections and the average actual reward. 
The team's expected reward for the perfect selections in 1000~Plays is denoted as $R^* = (P_{\mathrm{1st}} + P_{\mathrm{2nd}}) \times 1000$, where $P_{\mathrm{1st}}$ and $P_{\mathrm{2nd}}$ represent the hit probability of the best and second-best slot, respectively. 
The actual reward averaged over 2000 trials is denoted as $R$. 
We employ absolute regret $R^* - R$ as well as relative regret $(R^*-R)/R^*$. 
Lower regret generally signifies a better performance, but absolute regret depends strongly on the magnitude of $R^*$ as well. 

Using the evaluation metrics, we first assess the algorithm's performance for the following five settings, including the control condition addressed in Sec.~\ref{subsec:onetry}: $(P_{\mathrm{A}}, P_{\mathrm{B}}, P_{\mathrm{C}}) = (0.1, 0.9, 0.9)$, $(0.2, 0.8, 0.8)$, $(0.3, 0.7, 0.7)$, $(0.4, 0.6, 0.6)$, $(0.45, 0.55, 0.55)$. 
These cases involve symmetrical hit probabilities for the top two slots, simplifying the analysis of the system's behavior. 
The smaller the difference between the good and bad slots, the more trials are required for accurate inferences. 
Therefore, it is expected that the learning speed is slower with the narrower gap between $P_{\mathrm{A}}$ and $P_{\mathrm{B}} (= P_{\mathrm{C}})$ if the algorithm is reasonable. 

\begin{figure}[t]
\centering\includegraphics[width = 0.5\linewidth]{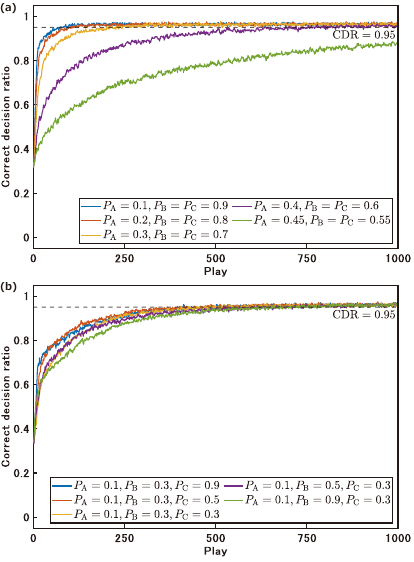}
\caption{Correct decision ratio (CDR) for 2000 trials. Various reward distributions are applied. The horizontal dotted line represents $\mathrm{CDR} = 0.95$. (a) Five different reward distributions with symmetric hit probabilities for the top two slots. (b) Five different reward distributions, including ones with asymmetry. }
\label{fig:cdr1}
\end{figure}
Figure~\ref{fig:cdr1}~(a) illustrates the evolution of CDR with regard to Play for five different reward distributions with symmetric hit probabilities for the top two slots. 
As shown in Fig.~\ref{fig:cdr1}~(a), CDR gradually increases with the number of Plays and converges to a value slightly greater than 0.95 as the number of Plays reaches 1000, except for $(P_{\mathrm{A}}, P_{\mathrm{B}}, P_{\mathrm{C}}) = (0.45, 0.55, 0.55)$ case. 
When the difference between reward probabilities becomes larger, it is easier to identify good and bad slots. 
Indeed, the proposed DCA method is able to quickly switch to exploitation in such cases, as one expects from an MAB algorithm.
Even with a challenging setting like the 0.45 versus 0.55 situation, CDR clearly keeps rising, indicating the effectiveness of our proposed DCA method. 

Next, to further discuss the system's characteristics, performance evaluations for the following five configurations are conducted: $(P_{\mathrm{A}}, P_{\mathrm{B}}, P_{\mathrm{C}}) = (0.1, 0.3, 0.9)$, $(0.1, 0.3, 0.5)$, $(0.1, 0.3, 0.3)$, $(0.1, 0.5, 0.3)$, $(0.1, 0.9, 0.3)$. 
Inconsistent behaviors can be observed when comparing such asymmetric configurations. 
They have commonly hit probabilities of 0.3 and 0.1 for the second and third-best slots. 
Also, these setups include pairs that are essentially the same but differ in the order of slot allocations, namely, $(P_{\mathrm{A}}, P_{\mathrm{B}}, P_{\mathrm{C}}) = (0.1, 0.3, 0.9)$ and $(0.1, 0.9, 0.3)$, and $(0.1, 0.3, 0.5)$ and $(0.1, 0.5, 0.3)$. 
Ideally, the system is expected to exhibit consistent behavior for these pairs due to their equivalence.  

Figure~\ref{fig:cdr1}~(b) shows the CDR changes against Play for the five reward distributions, including ones with asymmetric hit probabilities. 
These problem configurations also exhibit a steady increase in CDR, validating that the DCA method correctly operates in this regard.
However, when comparing the results for the pairs of equivalent problem setups, the outcomes shown in Fig.~\ref{fig:cdr1}~(b) are not desirable.
In settings $(P_{\mathrm{A}}, P_{\mathrm{B}}, P_{\mathrm{C}}) = (0.1, 0.9, 0.3)$ and $(0.1, 0.5, 0.3)$, the convergence of the CDR is slower than in settings $(P_{\mathrm{A}}, P_{\mathrm{B}}, P_{\mathrm{C}}) = (0.1, 0.3, 0.9)$ and $(0.1, 0.3, 0.5)$ despite their equivalent levels of difficulty. 
It is even slower compared to the setting $(P_{\mathrm{A}}, P_{\mathrm{B}}, P_{\mathrm{C}}) = (0.1, 0.3, 0.3)$, which is intuitively more challenging since the best slot is even harder to identify there. 
This behavior can be attributed to the asymmetric leader probabilities of lasers that mainly act as `laggards' due to the unidirectional coupling in the six-laser network. 
Once stuck in a wrong option, the ease of escaping from it can depend on the slight probability of the truly correct option being selected.  
Similar trends have been reported in the previous study on addressing the single-player multi-armed bandit problem using a unidirectional ring-connected laser network~\cite{Mihana2020}. 
While the previous work has shown that the gap in CDR curves for rearranged orders of hit probabilities of slot expands with more Plays, our proposed DCA is able to gradually diminish this difference, resulting in the convergence of CDR to nearly the same value independent of the ordering of reward probabilities, as shown in Fig.~\ref{fig:cdr1}~(b). 

\begin{table}[t]
\caption{Mean of correct decision ratio (CDR) over 1000~Plays, absolute (abs.) regret, and relative (rel.) regret in numerical simulations for various reward distributions. }
\label{tab:regret1}
\begin{center}
{
\footnotesize
     \begin{tabularx}{0.6\linewidth}{cCCC}
         \hline
         $(P_{\mathrm{A}}, P_{\mathrm{B}}, P_{\mathrm{C}})$ & Mean of CDR & Abs. regret & Rel. regret\\
         \hline
         $(0.1, 0.9, 0.9)$ & 0.956 & 35.1 & 0.020\\
         $(0.2, 0.8, 0.8)$ & 0.953 & 28.3 & 0.018\\
         $(0.3, 0.7, 0.7)$ & 0.942 & 22.8 & 0.016\\
         $(0.4, 0.6, 0.6)$ & 0.887 & 22.4 & 0.019\\
         $(0.45, 0.55, 0.55)$ & 0.752 & 24.6 & 0.022\\
         \hline
         $(0.1, 0.3, 0.9)$ & 0.919 & 27.5 & 0.023\\
         $(0.1, 0.3, 0.5)$ & 0.921 & 20.3 & 0.025\\
         $(0.1, 0.3, 0.3)$ & 0.908 & 18.2 & 0.030\\
         $(0.1, 0.5, 0.3)$ & 0.903 & 25.8 & 0.032\\
         $(0.1, 0.9, 0.3)$ & 0.892 & 38.2 & 0.032\\
         \hline
 \end{tabularx}
 }
\end{center}
\end{table}
Table~\ref{tab:regret1} displays CDR averaged over 1000~Plays, absolute regret, and relative regret in the numerical simulations for various reward distributions. 
For $(P_{\mathrm{A}}, P_{\mathrm{B}}, P_{\mathrm{C}}) = (0.3, 0.7, 0.7)$, while CDR remains consistently lower than for $(0.1, 0.9, 0.9)$ and $(0.2, 0.8, 0.8)$, absolute and relative regret are the lowest among the three configurations. 
This reflects the fact that the reduced gap between good and bad slots can soften the loss from incorrect decisions. 
Nevertheless, relative regret is increased for $(P_{\mathrm{A}}, P_{\mathrm{B}}, P_{\mathrm{C}}) = (0.4, 0.6, 0.6)$ and $(0.45, 0.55, 0.55)$ due to slow learning, as evidenced by the gradual increase in CDR. 
The lower five settings exhibit noticeably higher average CDR than $(P_{\mathrm{A}}, P_{\mathrm{B}}, P_{\mathrm{C}}) = (0.45, 0.55, 0.55)$. 
However, relative regret is worse because the difference in the expected rewards between the best and worst options is significant for these settings with asymmetric hit probabilities, especially for $(P_{\mathrm{A}}, P_{\mathrm{B}}, P_{\mathrm{C}}) = (0.1, 0.5, 0.3)$ and $(0.1, 0.9, 0.3)$, where the worst option is more frequently selected. 

Overall, the numerical simulations have demonstrated that learning of our proposed DCA method robustly succeeds across various configurations. 
Naturally, the algorithm performance possibly varies depending on the hyperparameters of DCA, whose dependence will be discussed in the following section. 

\subsection{\label{subsec:kappa} Effects of hyperparameters}
In this section, we examine how the hyperparameters of our proposed decentralized coupling adjustment (DCA), described by Eqs.~\eqref{eq:reltow1} and \eqref{eq:reltow3}, influence the decision-making system's performance. 
This analysis reveals requirements for an effective system to address the competitive multi-armed bandit (CMAB) problem, enabling us to enhance algorithm design. 
Hyperparameters of DCA are as follows: scaling factor $r_{\mathrm{step}}$, the lower and upper bound of coupling strength $\kappa_{\mathrm{low}}$ and $\kappa_{\mathrm{upp}}$. 
Also, there is flexibility in the definition of the initial attenuation rate $r_{\mathrm{ini}}$. 
Here, we highlight only $r_{\mathrm{step}}$ and $\kappa_{\mathrm{low}}$. 

First, we focus on scaling factor $r_{\mathrm{step}}$.
From Eq.~\eqref{eq:reltow3}, $r_{\mathrm{step}}$ corresponds to the proportion by which the attenuation rates $r_{1\sharp}$ and $r_{2\sharp}$ should be shifted from their baseline value $r_{\mathrm{ini}}$ in relation to $Q_{1,\mathrm{S}(\sharp)}$ and $Q_{2,\mathrm{S}(\sharp)}$ $(\mathrm{\sharp} = \mathrm{bl}, \mathrm{or}, \mathrm{ye})$. 
After sufficient exploration, $Q_{1,\mathrm{S}(\sharp)}$ and $Q_{2,\mathrm{S}(\sharp)}$ settle to certain values prescribed by the problem setting $(P_{\mathrm{A}}, P_{\mathrm{B}}, P_{\mathrm{C}})$. 
The resultant $r_{1\sharp}$ and $r_{2\sharp}$ also converge to specific values or reach the lower or upper bound, $r_{\mathrm{low}}$ or $r_{\mathrm{upp}}$.
The extent of exploitation is stronger when there is a significant difference between coupling strengths, and thus, scaling factor $r_{\mathrm{step}}$ basically determines the strength of exploitation after the exploration phase; exploitation can become insufficient when $r_{\mathrm{step}}$ is not large enough. 

\begin{figure}[t]
\centering\includegraphics[width = 0.5\linewidth]{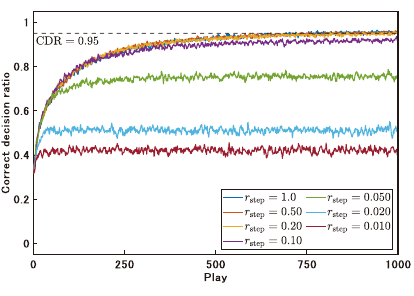}
\caption{Correct decision ratio (CDR) for 2000 trials. Seven different values of scaling factor $r_{\mathrm{step}}$ are applied for a setting $(P_{\mathrm{A}}, P_{\mathrm{B}}, P_{\mathrm{C}}) = (0.4, 0.6, 0.6)$. The horizontal dotted line represents $\mathrm{CDR} = 0.95$. }
\label{fig:cdr3}
\end{figure}
Figure~\ref{fig:cdr3} depicts the evolution of the correct decision ratio (CDR) as the number of Play progresses when applying the control condition $r_{\mathrm{step}} = 1.0$ or less than that. 
The hit probabilities of slot machines are set to $(P_{\mathrm{A}}, P_{\mathrm{B}}, P_{\mathrm{C}}) = (0.4, 0.6, 0.6)$. 
Although CDR curves exhibit similar initial rise rates, the values at which they converge vary depending on the value of $r_{\mathrm{step}}$. 
The DCA method yields an overly exploratory strategy with a low value of $r_{\mathrm{step}}$, and CDR does not get sufficiently close to 1.0 despite the sufficient number of Plays, as shown in Fig.~\ref{fig:cdr3}. 

Furthermore, the required minimum value of $r_{\mathrm{step}}$ also depends on the hit probabilities of slots since they affect the behavior of $Q_{1,\mathrm{S}(\sharp)}$ and $Q_{2,\mathrm{S}(\sharp)}$. 
If the difference between the good and bad slots is significant, i.e., the problem is not challenging, $Q_{1,\mathrm{S}(\sharp)}$ and $Q_{2,\mathrm{S}(\sharp)}$ converge to the larger absolute values, and exploitation naturally becomes stronger even if $r_{\mathrm{step}}$ is the same. 

\begin{figure}[t]
\centering\includegraphics[width = 0.5\linewidth]{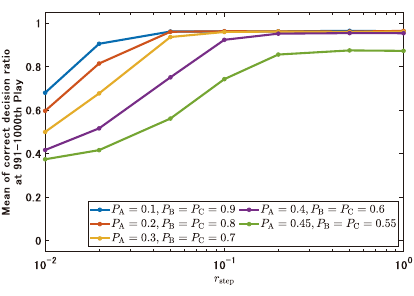}
\caption{The average of 991--1000th Play's correct decision ratio (CDR) for 2000 trials against scaling factor $r_{\mathrm{step}}$. Five different reward distributions are applied. }
\label{fig:cdr4}
\end{figure}
Figure~\ref{fig:cdr4} illustrates CDR averaged for the 991--1000th Plays for various $r_{\mathrm{step}}$ and $(P_{\mathrm{A}}, P_{\mathrm{B}}, P_{\mathrm{C}})$ configurations. 
As predicted, a decrease in $r_{\mathrm{step}}$ leads to lower CDR values across all problem scenarios due to insufficient exploitation. 
Moreover, CDR values drop earlier with more challenging problems, such as $(P_{\mathrm{A}}, P_{\mathrm{B}}, P_{\mathrm{C}}) = (0.4, 0.6, 0.6)$ or $(0.45, 0.55, 0.55)$. 

Next, we discuss the lower bound of coupling strengths $\kappa_{\mathrm{low}}$. 
$\kappa_{\mathrm{low}}$, which determines the lower limit of the attenuation rates $r_{\mathrm{low}} \equiv \sqrt{\kappa_{\mathrm{low}}/\kappa}$, accompanied by $\kappa_{\mathrm{upp}}$, which specifies the upper limit $r_{\mathrm{upp}} \equiv \sqrt{\kappa_{\mathrm{upp}}/\kappa}$, works as a parameter that determines the balance between exploration and exploitation; 
if the difference between $\kappa_{\mathrm{low}}$ and $\kappa_{\mathrm{upp}}$ is large, the probability of selecting the estimated-optimal option increases, potentially resulting in more effective exploitation. 
However, an excessively exploitative strategy can lead to incorrect estimations and getting stuck to ineffective options, especially in challenging problem settings. 
Ideally, $\kappa_{\mathrm{low}}$ and $\kappa_{\mathrm{upp}}$ should be explored in two dimensions, but our preliminary examination has revealed that the optimal value of $\kappa_{\mathrm{low}}$ shifts according to $\kappa_{\mathrm{upp}}$, and thus, we focus solely on $\kappa_{\mathrm{low}}$ for simplicity. 

\begin{figure}[t]
\centering\includegraphics[width = 0.5\linewidth]{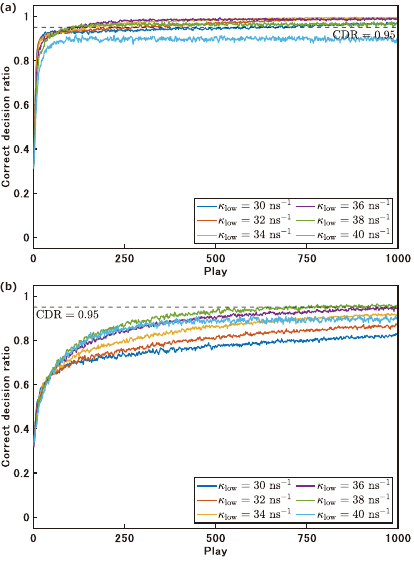}
\caption{Correct decision ratio (CDR) for 2000 trials. Six different values of the lower bound of coupling strength $\kappa_{\mathrm{low}}$ are applied. $\kappa_{\mathrm{low}}$ determines the lower limit of the attenuation rates $r_{\mathrm{low}}$. The horizontal dotted line represents $\mathrm{CDR} = 0.95$. (a) The problem setting of $(P_{\mathrm{A}}, P_{\mathrm{B}}, P_{\mathrm{C}}) = (0.2, 0.8, 0.8)$. (b) The problem setting of $(P_{\mathrm{A}}, P_{\mathrm{B}}, P_{\mathrm{C}}) = (0.4, 0.6, 0.6)$.}
\label{fig:cdr5}
\end{figure}
Figure~\ref{fig:cdr5} illustrates the changes in CDR with the accumulation of Plays for six different settings of the lower bound of coupling strength $\kappa_{\mathrm{low}}$, including the control condition, $\kappa_{\mathrm{low}} = \SI{38}{\per\nano\second}$. 
Generally, the lower value of $\kappa_{\mathrm{low}}$, that is, the larger the difference between $\kappa_{\mathrm{low}}$ and $\kappa_{\mathrm{upp}}$, the faster the initial increase in CDR. 
However, the subsequent behavior depends on the specific problem configuration. 
For the setting of $(P_{\mathrm{A}}, P_{\mathrm{B}}, P_{\mathrm{C}}) = (0.2, 0.8, 0.8)$ shown in Fig.~\ref{fig:cdr5}~(a), CDR converges to the highest value among the six configurations with $\kappa_{\mathrm{low}} = \SI{34}{\per\nano\second}$. 
In the $\kappa_{\mathrm{low}} = \SI{40}{\per\nano\second}$ setting, the CDR curve quickly saturates, and the eventual value is the lowest, resulting from an excessively exploratory strategy. 
On the other hand, for the challenging configuration $(P_{\mathrm{A}}, P_{\mathrm{B}}, P_{\mathrm{C}}) = (0.4, 0.6, 0.6)$, shown in Fig.~\ref{fig:cdr5}~(b), $\kappa_{\mathrm{low}} = \SI{38}{\per\nano\second}$ exhibits the highest CDR after 1000 Plays. 
With the configurations of $\kappa_{\mathrm{low}} = \SI{30}{\per\nano\second}$, $\SI{32}{\per\nano\second}$, and $\SI{34}{\per\nano\second}$, players overexploit early selections, leading to getting stuck on incorrect selections and significantly lower CDR values. 
Even so, CDR curves gradually keep recovering since the probability of selecting estimated suboptimal options is not zero. 
Thus, there is a possibility that the setting of $\kappa_{\mathrm{low}} = \SI{30}{\per\nano\second}$ can outperform that of $\kappa_{\mathrm{low}} = \SI{38}{\per\nano\second}$ in terms of CDR after further Plays. 

\begin{figure}[t]
\centering\includegraphics[width = 0.5\linewidth]{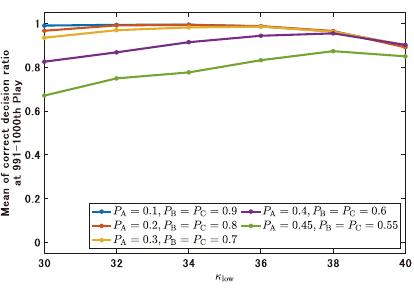}
\caption{The average of 991--1000th Play's correct decision ratio (CDR) for 2000 trials against scaling factor $\kappa_{\mathrm{low}}$. Five different reward distributions are applied. }
\label{fig:cdr6}
\end{figure}
Figure~\ref{fig:cdr6} shows the average of CDR at 991--1000th Play for various $\kappa_{\mathrm{low}}$ and $(P_{\mathrm{A}}, P_{\mathrm{B}}, P_{\mathrm{C}})$ configurations. 
An exploitative strategy with $\kappa_{\mathrm{low}} = \SI{30}{\per\nano\second}$ is effective for $(P_{\mathrm{A}}, P_{\mathrm{B}}, P_{\mathrm{C}}) = (0.1, 0.9, 0.9)$, whereas a deliberate approach with $\kappa_{\mathrm{low}} = \SI{38}{\per\nano\second}$ yields better results for $(P_{\mathrm{A}}, P_{\mathrm{B}}, P_{\mathrm{C}}) = (0.4, 0.6, 0.6)$ and $(0.45, 0.55, 0.55)$. 

Therefore, the appropriate $\kappa_{\mathrm{low}}$ and $\kappa_{\mathrm{upp}}$ greatly depend on the number of Plays and the acceptable lower limit of CDR or upper limit of regret. 
Additionally, there is potential for players to pick a better configuration depending on whether they can roughly estimate the hit probabilities of slot machines beforehand. 

\clearpage
\section{\label{sec:conclu} Conclusion}
We focused on collective decision-making utilizing a cluster-synchronized laser network~\cite{Ito2024, Kotoku2024} for addressing the competitive multi-armed bandit (CMAB) problem, one of the most fundamental configurations of multi-agent reinforcement learning (MARL). 
We proposed a six-laser network and decentralized coupling adjustment (DCA), extending the system in the earlier research and enabling it to resolve the two-player and three-slot CMAB problem with a photonic-based and straightforward learning algorithm. 
Our numerical simulations demonstrated that the laser network exhibits three-cluster synchronization as intended and that the learning algorithm effectively adapts the optical attenuation rate based on observed slot probabilities in a distributed manner. 
The average performance evaluation verified that learning steadily progressed regardless of various reward distributions and examined the effects of hyperparameter configurations of the DCA method. 

We treated the CMAB problem with two players and three slots as the minimal setting under the conditions requiring exploitation. 
Our preliminary analysis has implied that the problem with increased players and slots can be managed by further enlarging the laser network and increasing the number of clusters. 
However, the straightforward expansion leads to a prohibitive increase in the required number of lasers, indicating the need for new principles for selecting slots other than the one based on the leader-laggard relationship. 

We dealt with a time-invariant reward environment for simplicity. 
Time-varying environments can also be addressed by modifying the DCA method to estimate the hit probabilities of slot machines solely from the recent slot selection results or by introducing a memory parameter.

\section*{Funding}
This research was funded in part by the Japan Society for the Promotion of Science through Grant-in-Aid for Research Activity Start-up (22K21269), Grant-in-Aid for Early-Career Scientists (23K16961), and Transformative Research Areas (A) (JP22H05197).

\section*{Disclosures}
The authors declare no conflicts of interest.

\section*{Data Availability}
Data underlying the results presented in this paper are not publicly available at this time but may be obtained from the authors upon reasonable request.


%

\end{document}